\documentclass[a4paper, 10pt, conference]{ieeeconf}      

\IEEEoverridecommandlockouts                              

\usepackage{times}
\usepackage{epsfig}
\usepackage{graphicx,subfig}
\usepackage[tbtags]{amsmath}
\usepackage{amssymb}
\usepackage{float}
\usepackage{cite}
\usepackage[ruled,linesnumbered]{algorithm2e} 
\usepackage{algorithmicx}  
\usepackage{algpseudocode}  
\usepackage{CJK}
\usepackage{enumerate}
\usepackage{authblk}
\usepackage{stfloats}
\usepackage{mwe}
\usepackage{cuted}
\usepackage{capt-of}
\usepackage{caption}

\allowdisplaybreaks[3]

\usepackage{expl3}
\ExplSyntaxOn
\newcommand\latinabbrev[1]{
  \peek_meaning:NTF . {
    #1\@}%
  { \peek_catcode:NTF a {
      #1.\@ }%
    {#1.\@}}}
\ExplSyntaxOff

\usepackage[pagebackref=false,breaklinks=true,letterpaper=true,bookmarks=false,urlcolor=black, citecolor=black]{hyperref}
\hypersetup{hidelinks}

\title{\LARGE \bf
A General Framework of Learning Multi-Vehicle Interaction\\ Patterns from Videos
}

\author{Chengyuan Zhang$^{1}$, Jiacheng Zhu$^{2}$, Wenshuo Wang$^{2}$, Ding Zhao$^{2*}$
\thanks{*Corresponding author. E-mail: {\tt\small dingzhao@cmu.edu}}
\thanks{$^{1}$Chengyuan Zhang is with the Department of Automotive Engineering, Chongqing University, Chongqing 400044, China. 
}
\thanks{$^{2}$Jiacheng Zhu, Wenshuo Wang and Ding Zhao are with the Department of Mechanical Engineering, Carnegie Mellon University, Pittsburgh, PA 15213, USA.
}
}%

\begin{document}

\maketitle
\thispagestyle{empty}
\pagestyle{empty}

\begin{abstract}

Semantic learning and understanding of multi-vehicle interaction patterns in a cluttered driving environment are essential but challenging for autonomous vehicles to make proper decisions. This paper presents a general framework to gain insights into intricate multi-vehicle interaction patterns from bird's-eye view traffic videos. We adopt a Gaussian velocity field to describe the time-varying multi-vehicle interaction behaviors and then use deep autoencoders to learn associated latent representations for each temporal frame. Then, we utilize a hidden semi-Markov model with a hierarchical Dirichlet process as a prior to segment these sequential representations into granular components, also called traffic primitives, corresponding to interaction patterns. Experimental results demonstrate that our proposed framework can extract traffic primitives from videos, thus providing a semantic way to analyze multi-vehicle interaction patterns, even for cluttered driving scenarios that are far messier than human beings can cope with.

\end{abstract}

\begin{figure*}[t]
    \centering
    \includegraphics[width = 0.95\textwidth]{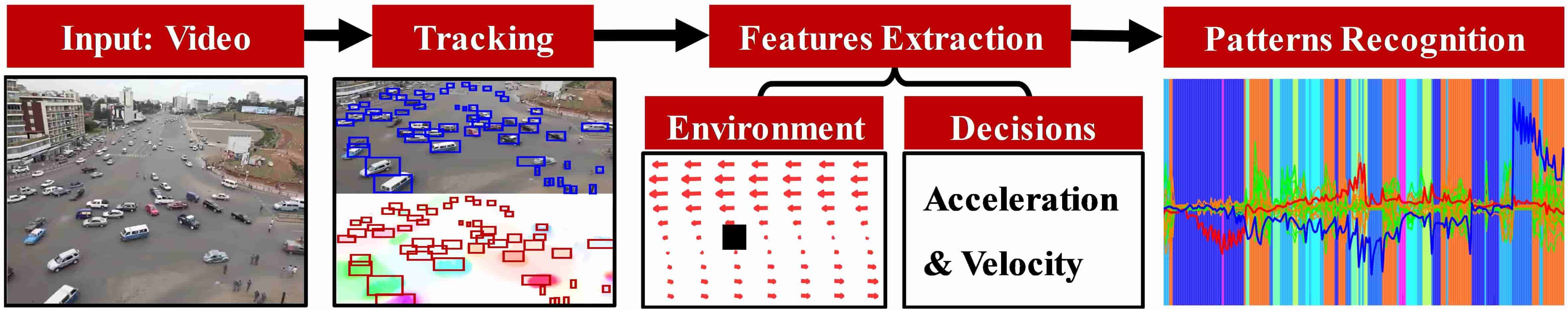}
    \caption{The proposed framework of extracting multi-vehicle interaction patterns from traffic videos with cluttered situations.}
    \label{fig1}
\end{figure*}

\section{INTRODUCTION}\label{sec1}


Semantic understanding of the dynamic interactions with their surroundings allows autonomous vehicles to make an accurate prediction in real-time, thus making proper decisions. It is essential but challenging to achieve such tasks because real-world driving scenarios are complex. Approaches with a specific simplified scenario or maneuver, such as lane-keeping, lane-changing, and merging on the highway \cite{gonzalez2017interaction,gindele2010probabilistic,li2018generic,sarkar2017trajectory}, have been proposed to study the interactions among multiple traffic participants. 
However, real-world traffic scenarios, such as intersections, are far more complicated. Here we propose a general framework to investigate how experienced human drivers interact with surrounding vehicles. This framework is capable of modeling interaction and learning interactive patterns from cluttered traffic scenarios. We anticipate this framework to be a starting point for analyzing of more sophisticated real-world traffic scenarios, including  reliable environment recognition, efficient scene understanding, and tractable safety evaluation.


The recorded traffic videos in a specific region-of-interest (ROI) contain informative multi-vehicle interaction patterns. The Next-Generation Simulation (NGSIM) \cite{NGSIM} is one of the most widely used datasets, which is recorded on the highway with explicit traffic rules, e.g., clear lane markers and well-controlled traffic lights. Some public datasets are also provided in \cite{dataset1,wang2007unsupervised,hu2006system}. In these datasets, the traffic signals may provide simple traffic patterns. However, they may not be extended into more cluttered traffic scenarios without traffic lights and lane markers. In order to study the interaction patterns in the cluttered urban traffic scenarios, we analyzed videos recorded over an unsignalized junction. In this paper, we provide a general framework capable of analyzing more complicated interaction patterns in the Meskel Square Dataset, without any traffic lights and clear lane markers. This dataset comprehensively reveals how human drivers interact with and react to each other in cluttered intersection scenarios without any external control, such as traffic lights.

Many approaches have been used to model interactions within multiple traffic participants. The model-based methods provide a straightforward understanding but maybe only suitable for limited scenarios \cite{klingelschmitt2016probabilistic}. The dynamic Bayesian networks\cite{gindele2010probabilistic} and deep neural networks\cite{sarkar2017trajectory} are powerful in inferring hidden states of traffic behavior, but the states should be in a fixed space. The inverse reinforcement learning (IRL) can model the decision-making process in multi-vehicle interaction behaviors \cite{gonzalez2017interaction}, but it may require prior knowledge to design appropriate reward functions. Besides, the IRL is a deterministic approach and could only pursue optimal motions. Therefore, these approaches cannot address situations with a time-varying number of interacting vehicles or complicated scenarios without explicit traffic rules.


Towards this end, our proposed general framework (Fig.\ref{fig1}) is capable of analyzing multi-vehicle interaction behaviors from complicated traffic videos without any traffic controls, e.g., traffic lights and lane markers, and we validate this framework on the Meskel Square Dataset. In this framework, we first extract the position and velocity of vehicles from videos by implementing a detection-based tracking algorithm that integrates You Only Look Once (\emph{YOLOv3}) \cite{redmon2018yolov3} with optical flow. However, the number of vehicles in each frame of the videos is changing, which lays a top challenge to define the interaction patterns. Then, to overcome this issue, we utilize a Gaussian velocity field to represent the {\it spatial} interaction patterns in each frame and apply a deep autoencoder to encode the high-dimensional representatives into a low-dimensional vectored latent space. Finally, we use Bayesian nonparametric learning to cluster the {\it temporal} interaction patterns with the encoded latent features. The main contributions are as follows:
\begin{enumerate}[1)]
    \item Providing a general framework to extract interaction patterns from traffic videos, which can be applied to any multi-agent interactions in videos.
    \item Introducing a Gaussian velocity field to describe interaction patterns in the spatial space, whose dimensions are invariant to the number of vehicles in the ROI, and thus can adapt to various scenarios from two vehicles \cite{wang2018clustering, wang2018understanding} to multiple vehicles.
    \item Employing Bayesian nonparametrics to learn interaction patterns in the temporal space automatically. 
\end{enumerate}
In addition, this developed framework can also provide a potential way to get high-quality data sources from traffic videos for multi-agent interaction analysis.

The remainder of this paper is structured as follows. Section \ref{sec2} introduces the essentials of our proposed framework. Section \ref{sec3} presents the details of motion extraction from videos. Section \ref{sec4} analyzes experimental results. Section \ref{sec5} makes further conclusion and discussion.

\section{Multi-Vehicle Interaction Modeling}\label{sec2}
In this section, we will introduce the Gaussian velocity field for describing the multi-vehicle interactions in each video frame, the autoencoder for learning representatives of the Gaussian velocity field, and Bayesian nonparametrics for clustering interaction patterns.
\subsection{Gaussian Velocity Field}
In most ROI of traffic, e.g., intersections, the number of vehicles randomly vary over time, which yields a time-varying dimension problem. 
There are two ways to deal with this problem: 1) Fixing the dimension of the extracted features that describe multi-vehicle interactions, or 2) Fixing the number of surrounding vehicles as a constant. Here we learn the fixed-dimension features with an assumption that the ego vehicle can perceive its nearby vehicles in terms of the position and relative velocity. Modeling motion patterns as velocity fields rather than trajectories is a robust method to group vehicles that share similar motion characteristics. Therefore, we employed a velocity field\cite{joseph2011bayesian, aoude2013probabilistically} to model the interaction patterns of surrounding vehicles with respect to the ego vehicle.

We model velocity field based on Gaussian process according to Joseph et al.\cite{joseph2011bayesian}. Given the observed locations of $M$ vehicles, $\{x_i, y_i\}_{i=1}^M$, their trajectory derivatives $\{\Delta x_{i}, \Delta y_{i}\}_{i=1}^M$ are jointly distributed according to a Gaussian distribution with mean velocity $\{\mu _{\Delta x/\Delta t}(x_i, y_i), \mu _{\Delta y/\Delta t}(x_i, y_i)\}_{i=1}^M$ and covariance $\{\Sigma_x, \Sigma_y\}$, where $\Sigma_x$ and $\Sigma_y$ denote the covariance of the $x$ and $y$ direction. In the $x$ direction, the components of $\Sigma_x$ can be expressed as $\Sigma _{x_{ij}}=K_{x}(x_i, y_i, x_j, y_j)$, where $K_{x}$ is the standard squared exponential covariance functions
\begin{equation}
\begin{split}
K_x(x_1, y_1, x_2, y_2) & = A\exp\left({-\frac{(x_1-x_2)^2}{2\sigma ^2_x} - \frac{(y_1-y_2)^2}{2\sigma ^2_y}}\right)\\
\end{split}
\end{equation}
and $A$ is the amplification factor, $\sigma_x$ and $\sigma_y$ are the length-scale factors governing the impact on each vehicle with different ranges.
\begin{figure}[t]
    \centering
    \includegraphics[width=0.9\linewidth]{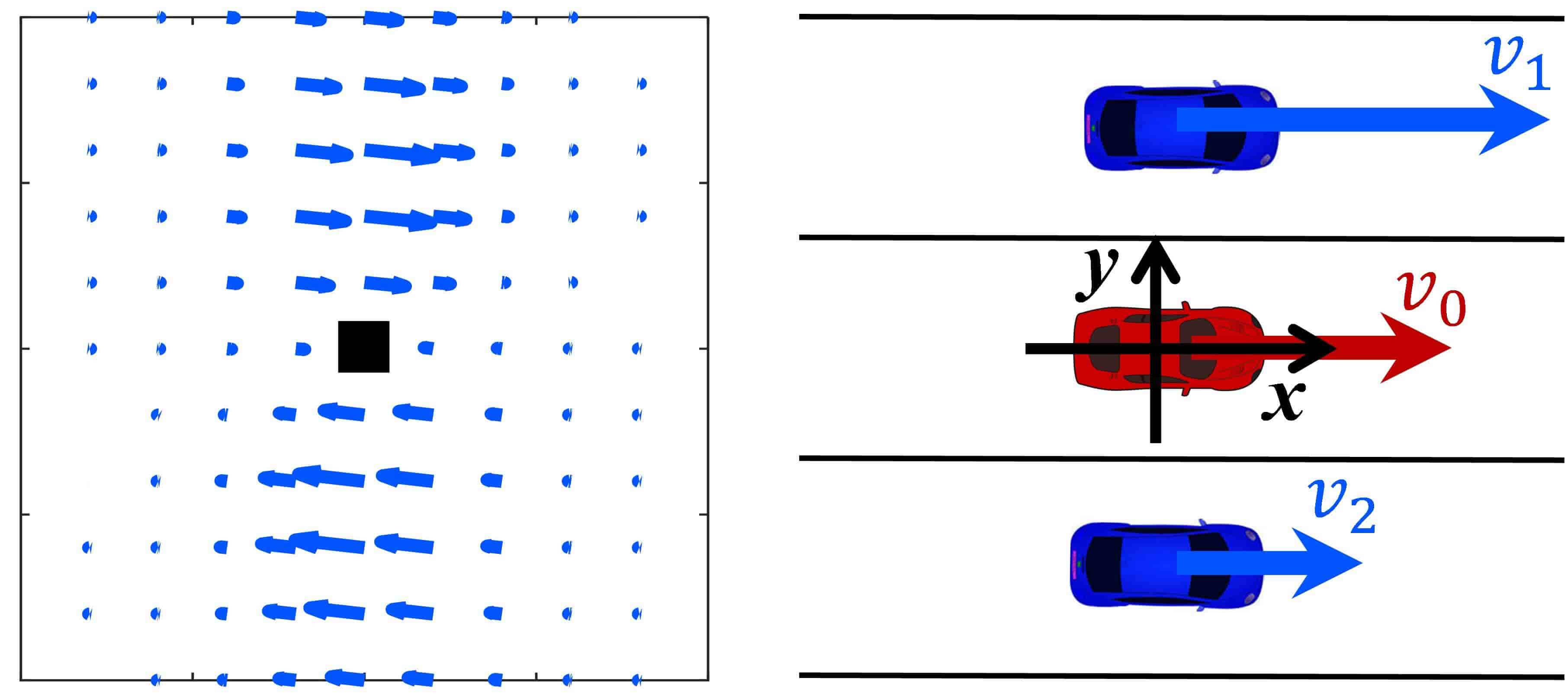}
    \caption{The Gaussian velocity field of multi-vehicle scenarios. 
    }
    \label{fig4}
\end{figure}
Given a new position $(x, y)$, the predictive distribution over the trajectory derivative $\Delta x / \Delta t$ via a Gaussian process can be computed by
\begin{align}
\mu _{\frac{\Delta x}{\Delta t}} & =K_x(x, y, \mathbf{X}, \mathbf{Y}) K_x(\mathbf{X}, \mathbf{Y}, \mathbf{X}, \mathbf{Y}) ^{-1}\frac{\Delta \mathbf{X}}{\Delta t}\\
\sigma _{\frac{\Delta x}{\Delta t}}^2 & = K_x(x, y, \mathbf{X}, \mathbf{Y}) K_x(\mathbf{X}, \mathbf{Y}, \mathbf{X}, \mathbf{Y}) ^{-1}K_x(\mathbf{X}, \mathbf{Y}, x, y)
\end{align}
where $\mathbf{X}$ and $\mathbf{Y}$ denote the positions of all surrounding vehicles at time $t$. The calculation of $\Delta y / \Delta t$ is similar to the procedure above by using $K_y$. Figure \ref{fig4} illustrates the Gaussian velocity field schematics with the relative velocities $v_1-v_0$ and $v_2-v_0$.

\subsection{Autoencoders}
An autoencoder is composed of an encoder $h=f(x)$ and a decoder $r=g(h)$. It reconstructs its inputs and sets the target values of output $\hat{x}=g(f(x))$ to be equal to $x$, where $f(\cdot )$ and $g(\cdot)$ are the activation functions. 
Then a deep autoencoder is formed through hierarchically stacking multiple autoencoders by treating the latent features $h$ in the middle hidden layer of $i^{th}$-autoencoder as the input of $(i+1)^{th}$-autoencoder.

\subsection{Traffic Primitives}
\subsubsection{Definition}

We define traffic primitives as the fundamental building blocks of multi-vehicle interactions that shape complicated scenarios. More specifically, traffic primitives are regarded as the non-overlapped segments of the time-series traffic observations with respect to interaction patterns. In this way, each traffic primitive represents an essential interaction behavior, and various complex scenarios are composed of several primitives. 

\subsubsection{Traffic Primitive Extraction}

The problem of multi-vehicle interaction is complicated with few studies have been done on this topic, thus setting a reasonable number of patterns is a tricky problem. We utilize Bayesian nonparametrics to learn the traffic primitives automatically, which does not require setting the number of traffic primitives initially. The Bayesian nonparametric model is a Bayesian model on an infinite-dimensional space which assumes the number of mixture model components increases with getting more observations\cite{orbanz2011bayesian}, in contrast to the approaches such as $k$-means in which the cluster number is predefined. In our research, the number of interaction patterns is formulated by a hierarchical Dirichlet process (HDP) which can optimize the number of interaction patterns. The relationship of sequential frames in the recorded video is formulated by a hidden semi-Markov model (HSMM). In turn, the combination of HDP and HSMM, called HDP-HSMM, can automatically segment the traffic time series into segments, called traffic primitives. In what follows, we will introduce the basic theoretical concepts of HDP-HSMM.

\begin{figure}[t]
    \centering
    \includegraphics[width=0.60\linewidth]{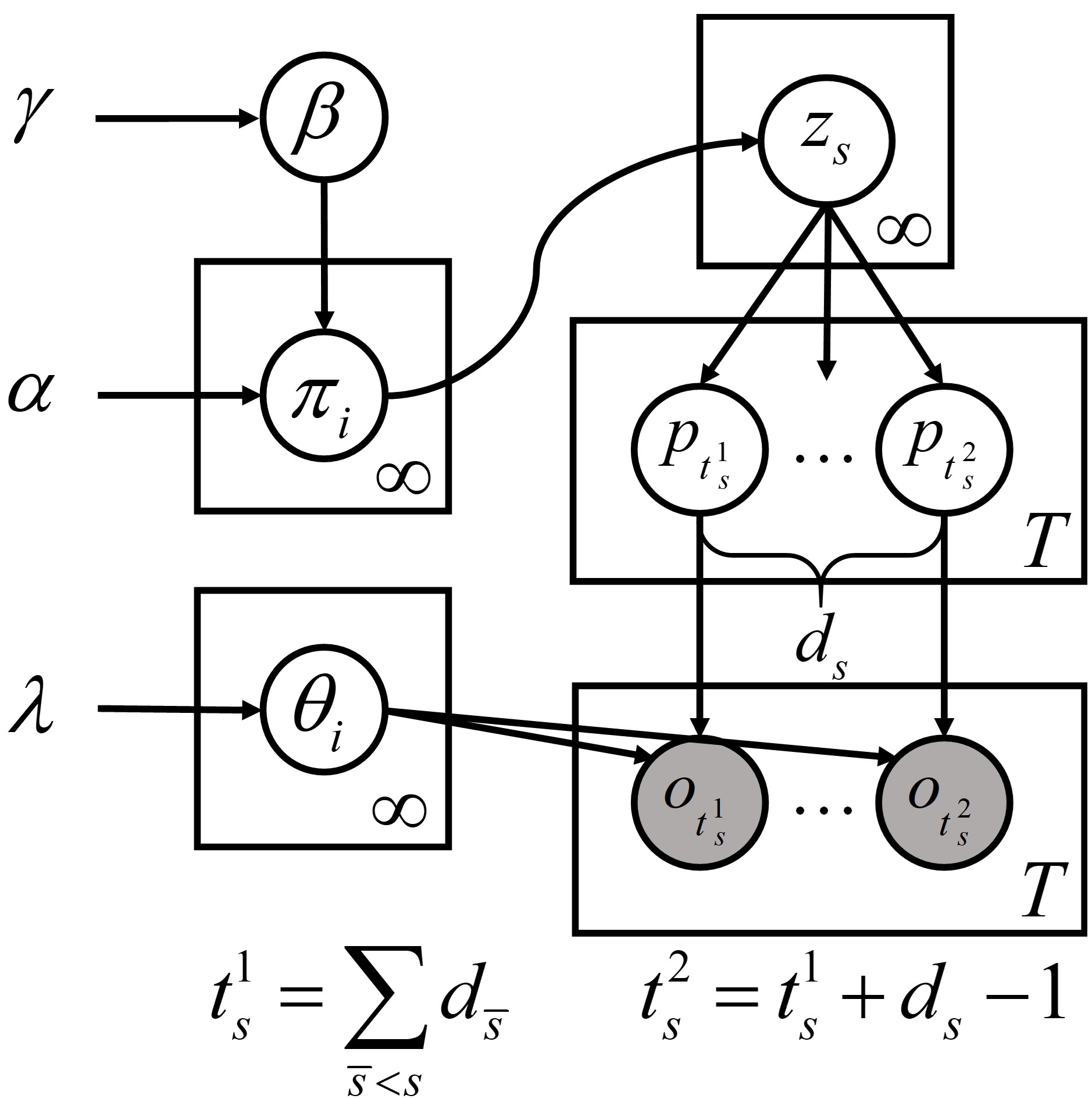}
    \caption{Graphical model of HDP-HSMM.}
    \label{fig6}
\end{figure}

The states transition process of interaction among multiple vehicles can be modeled as a probabilistic inferential process. 
For each state, a random state duration time is drawn from some state-specific distribution $g(\omega_s)$ which is set as Poisson prior. The hidden state of HSMM $p_t$ represents the traffic primitive and $o_t$ is the observed data at time $t$ \cite{johnson2013bayesian}. The duration of the primitive $d_t$ is a random variable, which is entered at time $t$, and $p(d_t|p_{t}=s_i)$ is the probability mass function. The HSMM can be interpreted as a Markov chain (without self-transitions) on a finite primitive set $P=\{s_i\}_{i=1}^S$. The transition probability from primitive $s_i$ to $s_j$ is defined as $\pi = \{\pi_{i, j(i\neq j)}\}_{i, j=1}^N$, where $N$ is the size of primitives set $P$. The possibility of observations $o_t$ given current primitive and the emission parameter $\theta_{p_t}$ of mode $s_i$ is defined as $p(o_t|p_t,\theta_{p_t})$. Thus, we can describe HSMM as
\begin{align}
p_t|p_{t-1} & \sim \pi_{p_{t-1}} \tag{4a}\\
d_t & \sim g(\omega_s)\tag{4b}\\
o_t|p_t,d_t & \sim F(\theta_{p_t},d_t)\tag{4c}
\end{align}
where $F(\cdot)$ is the emission function. The HDP can be formed by stick-breaking construction as
\begin{align}
\beta|\gamma &\sim {\rm GEM}(\gamma)\tag{5a}\\
\pi_j|\alpha,\beta &\sim {\rm DP}(\alpha,\beta)\tag{5b}\\
\theta_k|H &\sim H\tag{5c}\\
z_{ji}|\pi_j &\sim \pi_j\tag{5d}\\
o_{ji}|\{\theta_k\}_{k=1}^{\infty} &\sim F(\theta_{z_{ji}})\tag{5e}
\end{align}
where $z$ denotes the latent variables. By placing an HDP prior over the infinite transition matrices of HSMM, a robust architecture HDP-HSMM\cite{johnson2014bayesian} (Fig.\ref{fig6}) can be obtained
\begin{align}
\beta|\gamma &\sim {\rm GEM}(\gamma)\tag{6a}\\
\pi_j|\alpha,\beta &\sim {\rm DP}(\alpha,\beta), i=1,2,...\tag{6b}\\
(\theta_i,\omega_i)&\sim H\times G, i=1,2,...\tag{6c}\\
z_s &\sim \bar{\pi}_{z_{s-1}}, s=1,2,...\tag{6d}\\
d_s &\sim g(\omega_{z_s}), s=1,2,...\tag{6e}\\
p_{t_s^1:t_s^2}&=z_s\tag{6f}\\
o_{t_s^1:t_s^2}&\sim F(\theta_{p_t})\tag{6g}
\end{align}
where $t_s^1=\sum_{\bar{s}<s}d_{\bar{s}}$, $t_s^2=t_s^1+d_s-1$, and $\bar{\pi}_i=\frac{\pi_{ij}}{1-\pi_{ii}}(1-\delta_{ij})$ is added to eliminate self-transitions in the sequence $z_s$.

\subsubsection{Learning Procedure}

We adopt a weak-limit Gibbs sampling algorithm for HDP-HSMM\cite{johnson2013bayesian}; the duration variables are drawn from Poisson distribution. The observations are generated from a Gaussian model $\theta_i=[\mu_i,\Sigma_i]$, we take $\mu_i=0$ according to \cite{hamada2016modeling}. The hyperparameters $\alpha$ and $\gamma$ are drawn from a gamma prior, and the hyperparameters for $\theta$ are determined by an Inverse-Wishart (IW) prior\cite{wang2017driving}.

\section{Experimental Setup for Data Collection}\label{sec3}

In this section, to test the potential of the method capability, we will introduce a cluttered scenario, the Meskel Square Dataset (see \url{https://youtu.be/UEIn8GJIg0E}), which may never happen in the regular traffic. We will present a detection-based method to track dense moving objects in this low-resolution video. Then we will transform the tracked bounding boxes into the corresponding bird's eye vision. 

\subsection{Object Tracking}\label{subsec}
In order to extract the position and velocity of vehicles, two key procedures are necessary: 1) detection -- recognizing objects and tagging them with bounding boxes in each frame, and 2) tracking -- matching the same objects in successive frames. Many approaches are established on the optical flow as it can capture motions and segment objects \cite{dalal2006human}. However, the capacity of optical flow to detect static objects is limited. A powerful toolbox -- \emph{YOLOv3} \cite{redmon2018yolov3} -- has been developed based on convolutional neural networks, which we use in our research.

Given a set of bounding boxes without ID, a straightforward tracking way is to match the same bounding boxes between adjacent frames $t$ and $t+1$ using similarity measurements, e.g., Euclidean distance and overlapping ratio. The displacement of the same object between consecutive frames should be very small if the speed of objects is sufficiently low, implying that the point $p_{m, t}$ in the frame $t$ has a high possibility to match with the closest point $p_{n, t+1}$ in the frame $t+1$. Here we use the Euclidean Distance (ED) to gauge the similarity.

\begin{figure}[!t]
    \centering
    \subfloat[]{\label{fig2l}
    \includegraphics[width=0.3\linewidth]{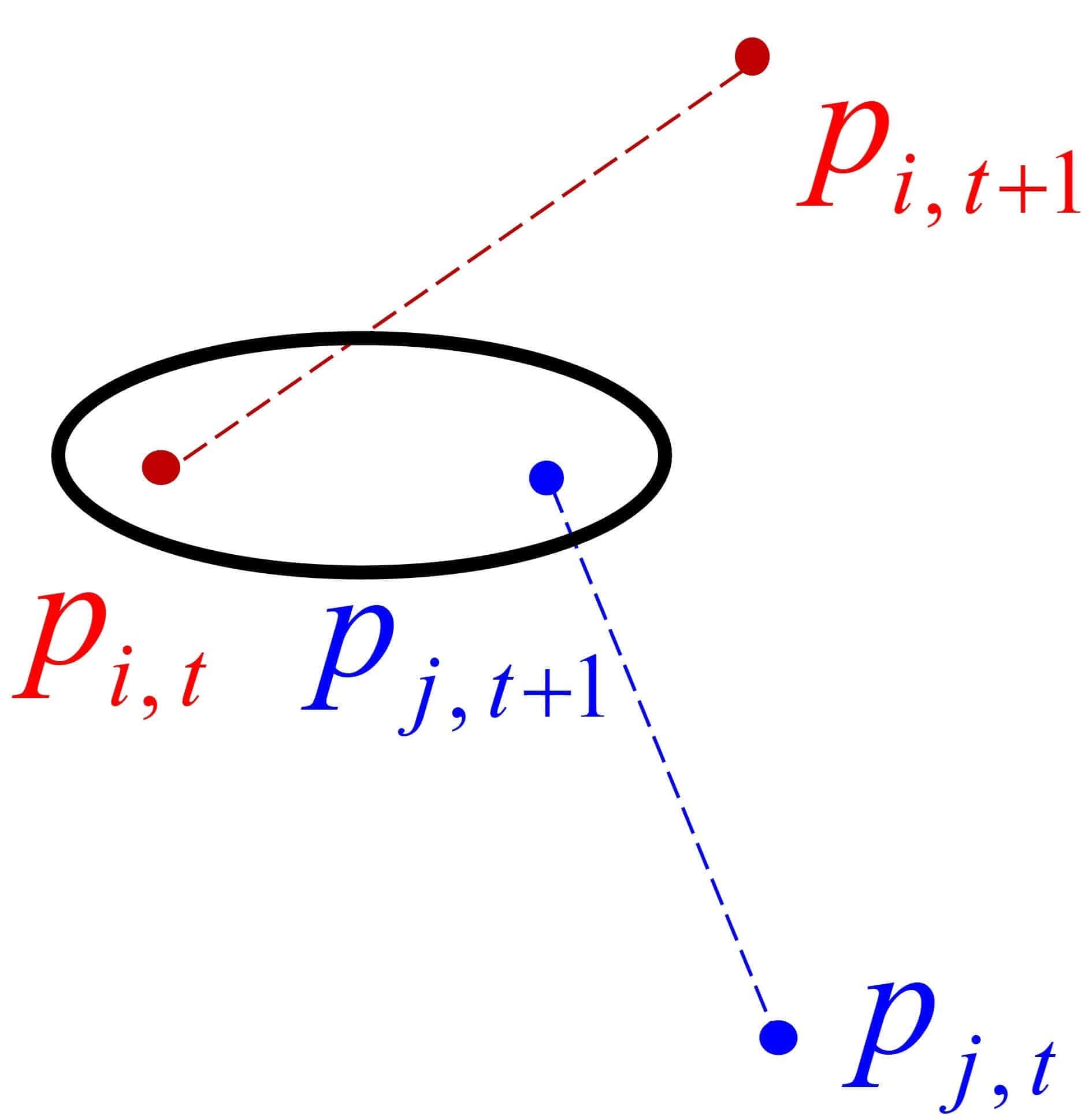}}
    \subfloat[]{\label{fig2m}
    \includegraphics[width=0.35\linewidth]{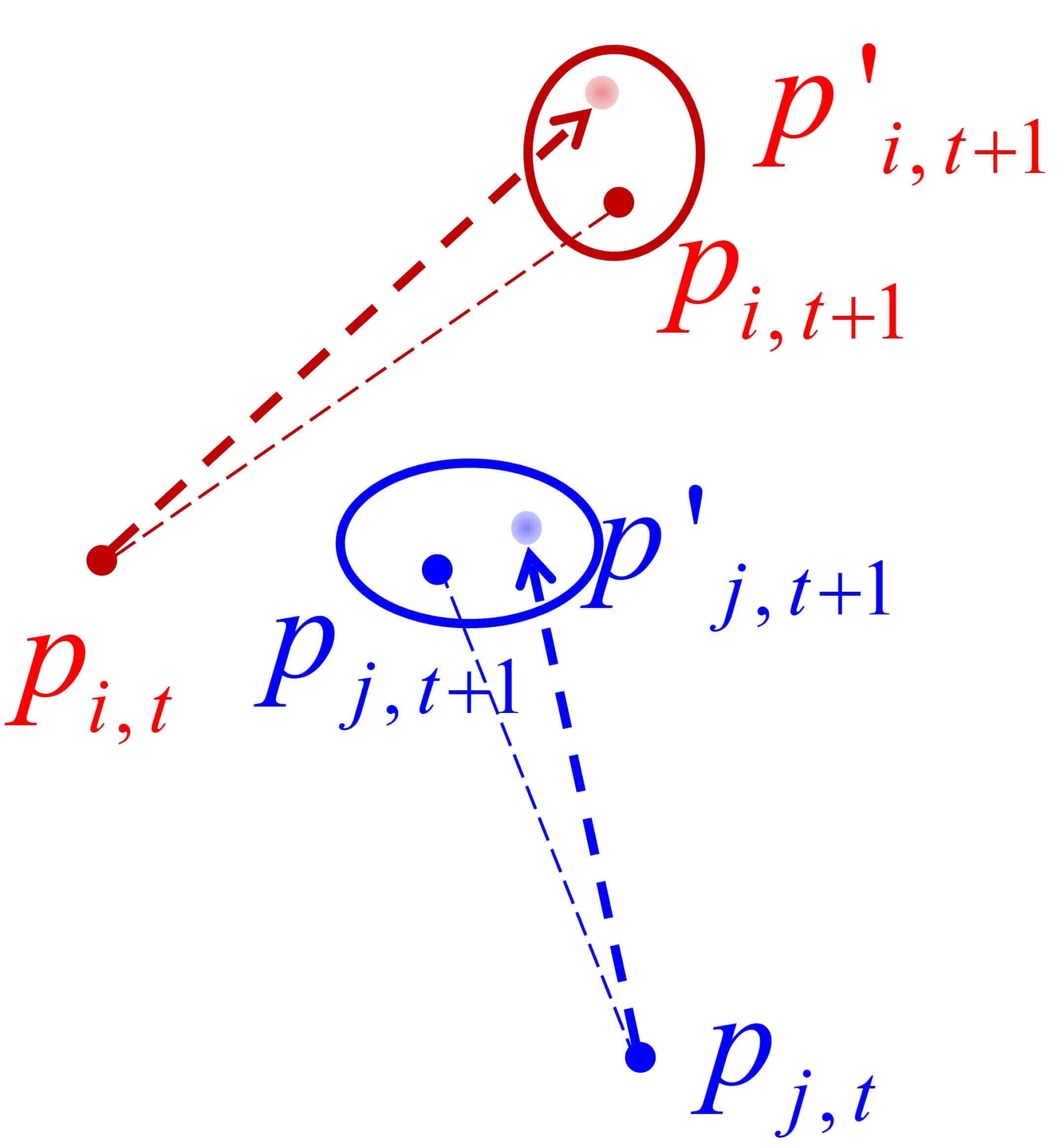}}
    \caption{The objects matching method of (a) directly minimizing ED (b) with movement prediction.}
    \label{fig2}
\end{figure}

Directly minimizing the ED may lead to incorrect results in some situations, for example, when the points in the frame $t$ are close to each other, e.g., $p_{i, t}$ is close to $p_{j, t}$ in Fig.\ref{fig2}\subref{fig2l}, it would match $p_{i, t}$ with $p_{j, t+1}$ incorrectly. To overcome this limitation, we have developed a matching algorithm based on movement prediction. The optical flow method which calculated the velocities of moving brightness patterns provides the trends of objects that can be used to predict the approximated upcoming position for each point. Therefore, the tracking method in Fig.\ref{fig2}\subref{fig2m} with the prediction of optical flow can match $p'_{i, t+1}$ (the predicted position of the object $i$ in the frame $t+1$) with $p_{i, t+1}$ correctly. 
In addition, the states and duration of each vehicle ID are also considered. An object is noted as inactive if no bounding box in the next frame was matched, and we do not update this ID once the duration of the inactive state is longer than a predefined threshold.

\subsection{Bird's-Eye View}

\begin{figure}[t]
    \centering
    \subfloat[]{\label{fig3l}
    \includegraphics[width=0.64\linewidth]{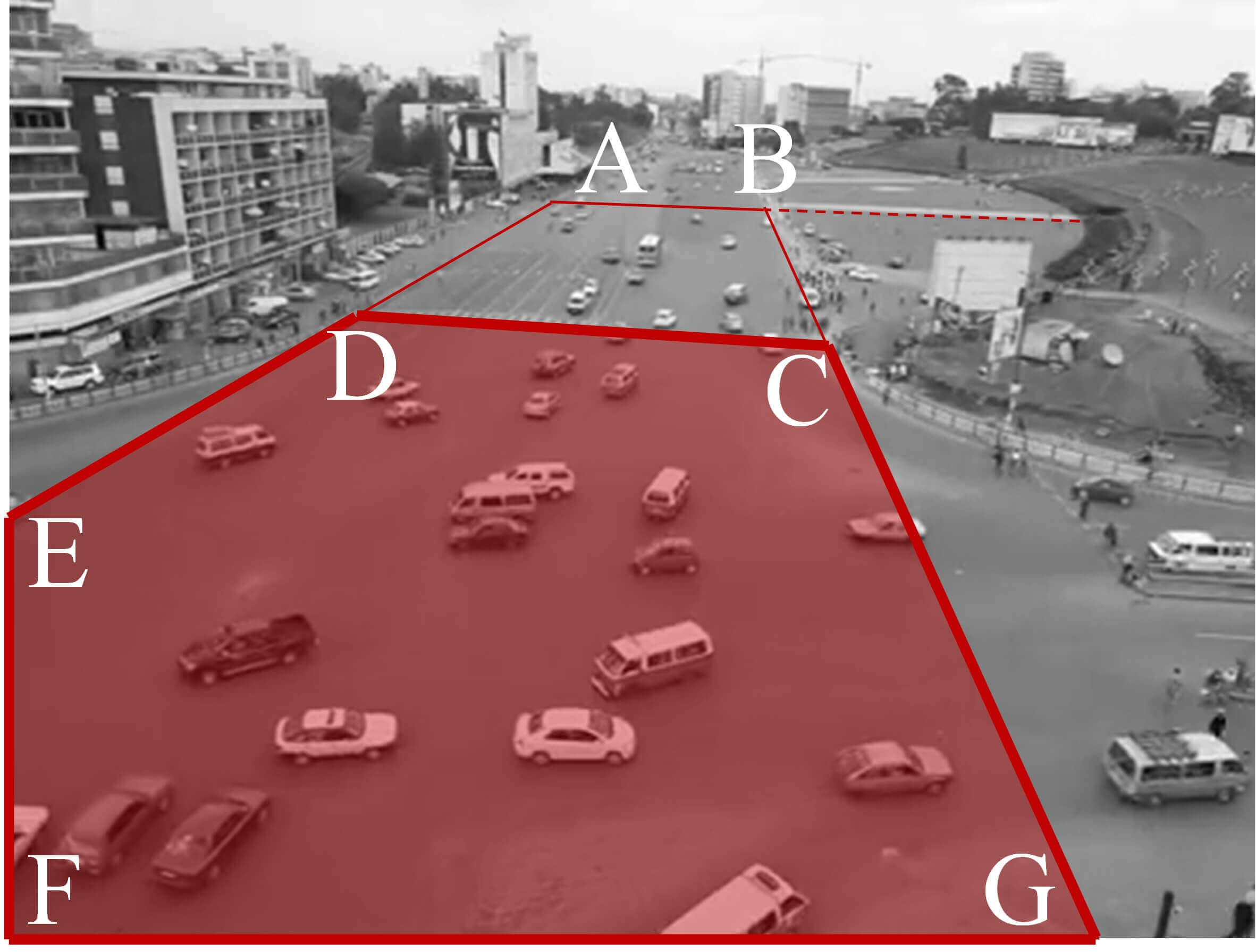}}
    \subfloat[]{\label{fig3r}
    \includegraphics[width=0.32\linewidth]{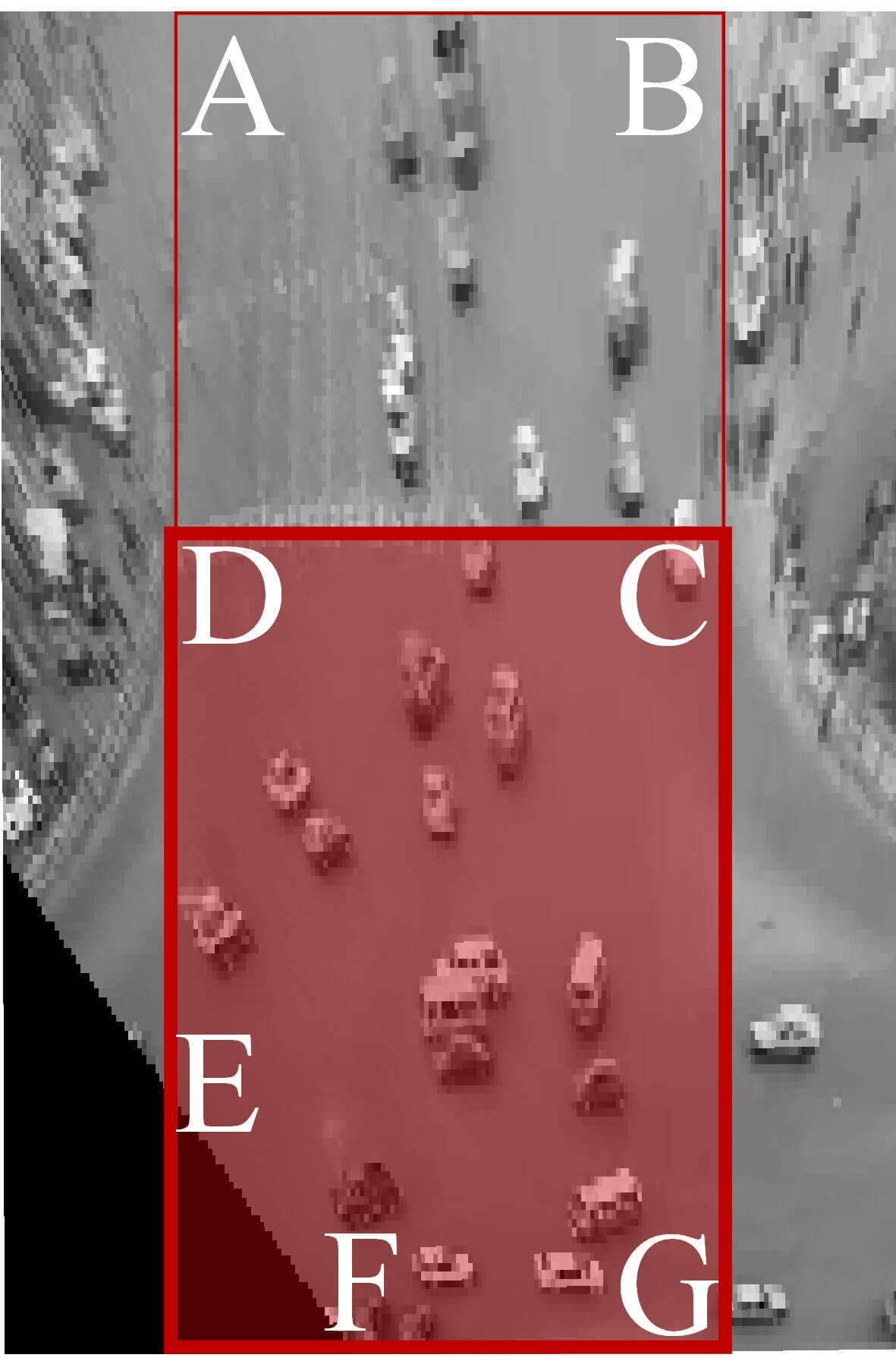}}
    \caption{A bird's-eye view perspective transformation with the line of the curb and crossing lines: (a) original frame, (b) top-down view.}
    \label{fig3}
\end{figure}

\begin{figure}[!t]
    \centering
    \subfloat[]{\label{fig10a}
    \includegraphics[width=0.25\linewidth]{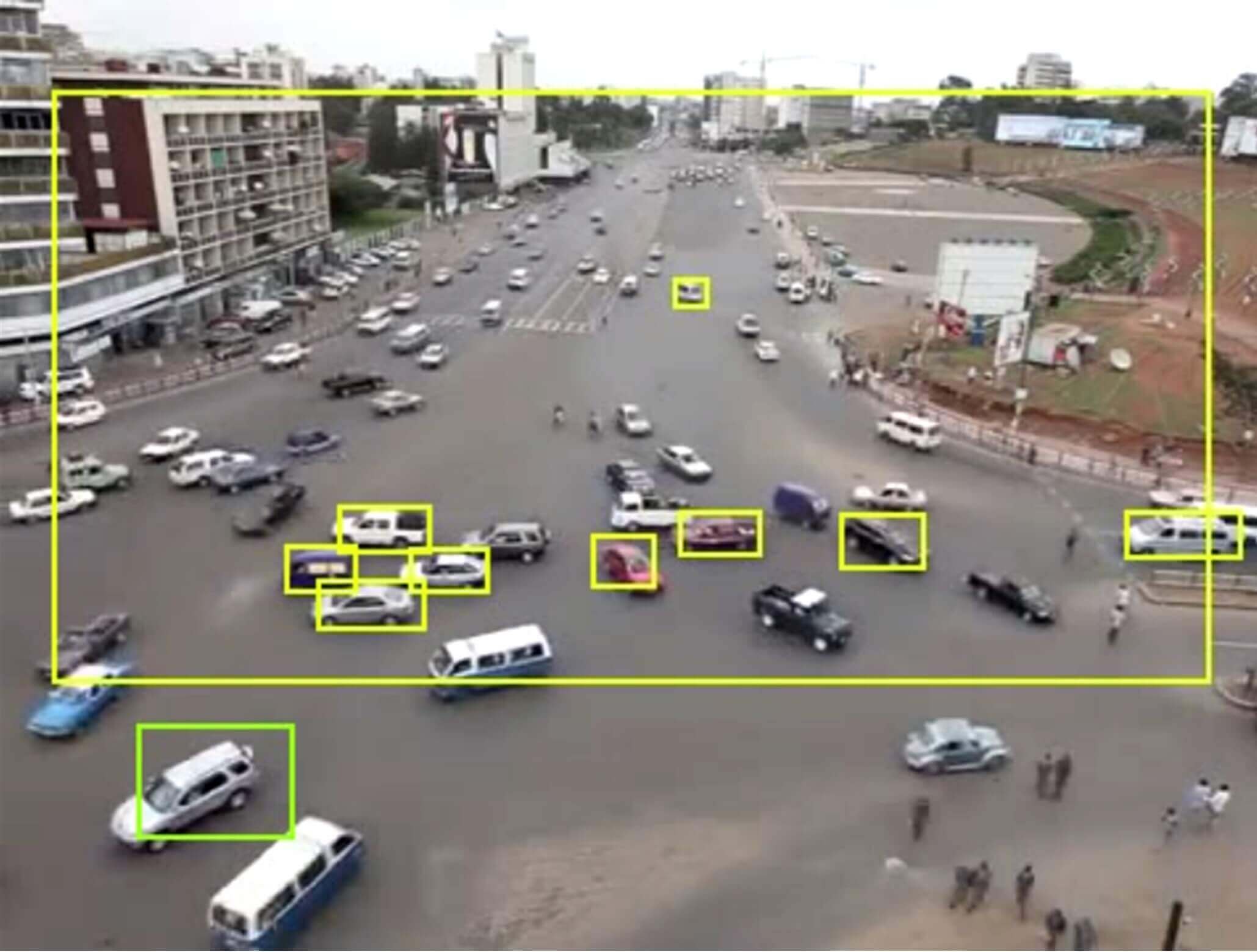}}
    \subfloat[]{\label{fig10b}
    \includegraphics[width=0.25\linewidth]{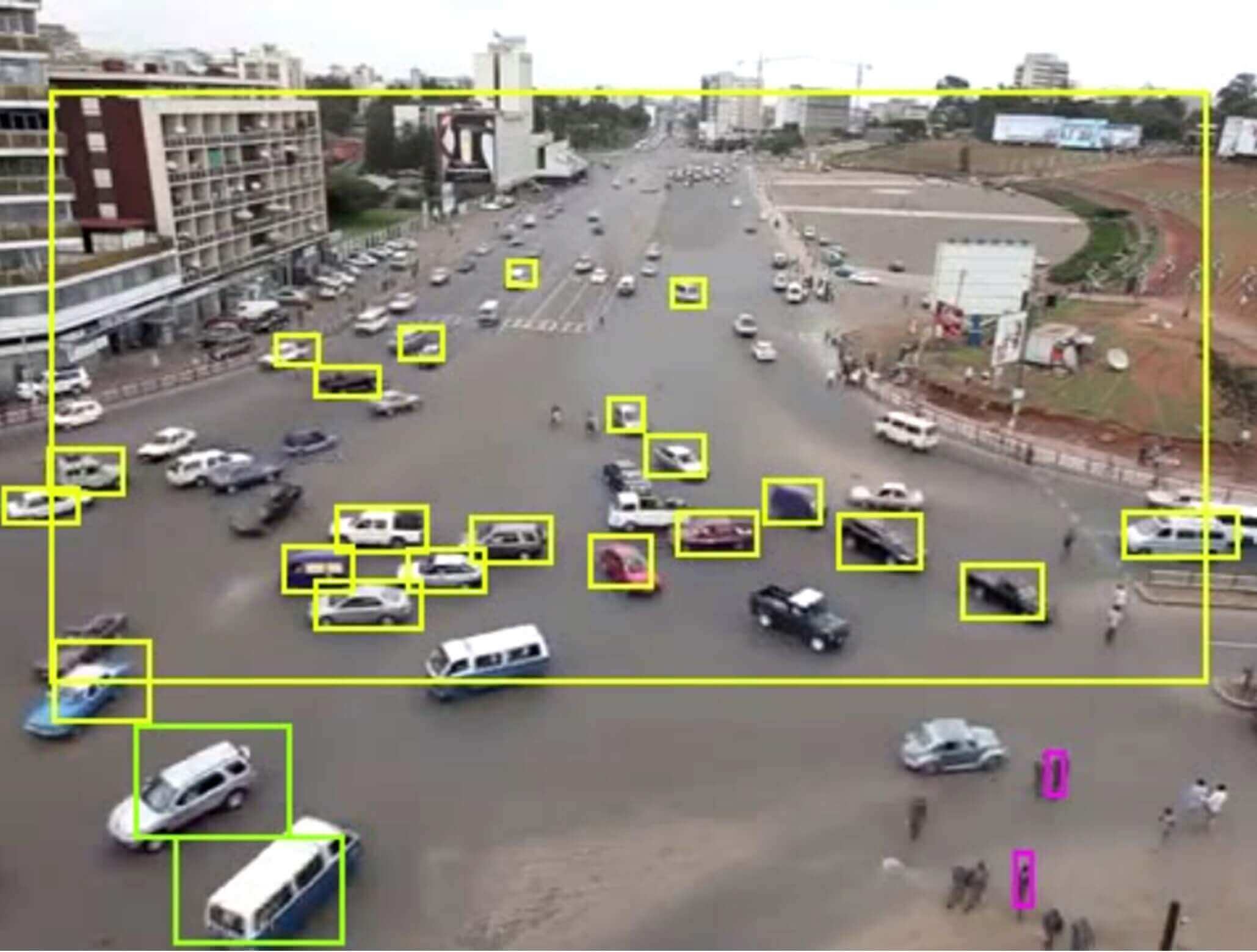}}
    \subfloat[]{\label{fig10c}
    \includegraphics[width=0.25\linewidth]{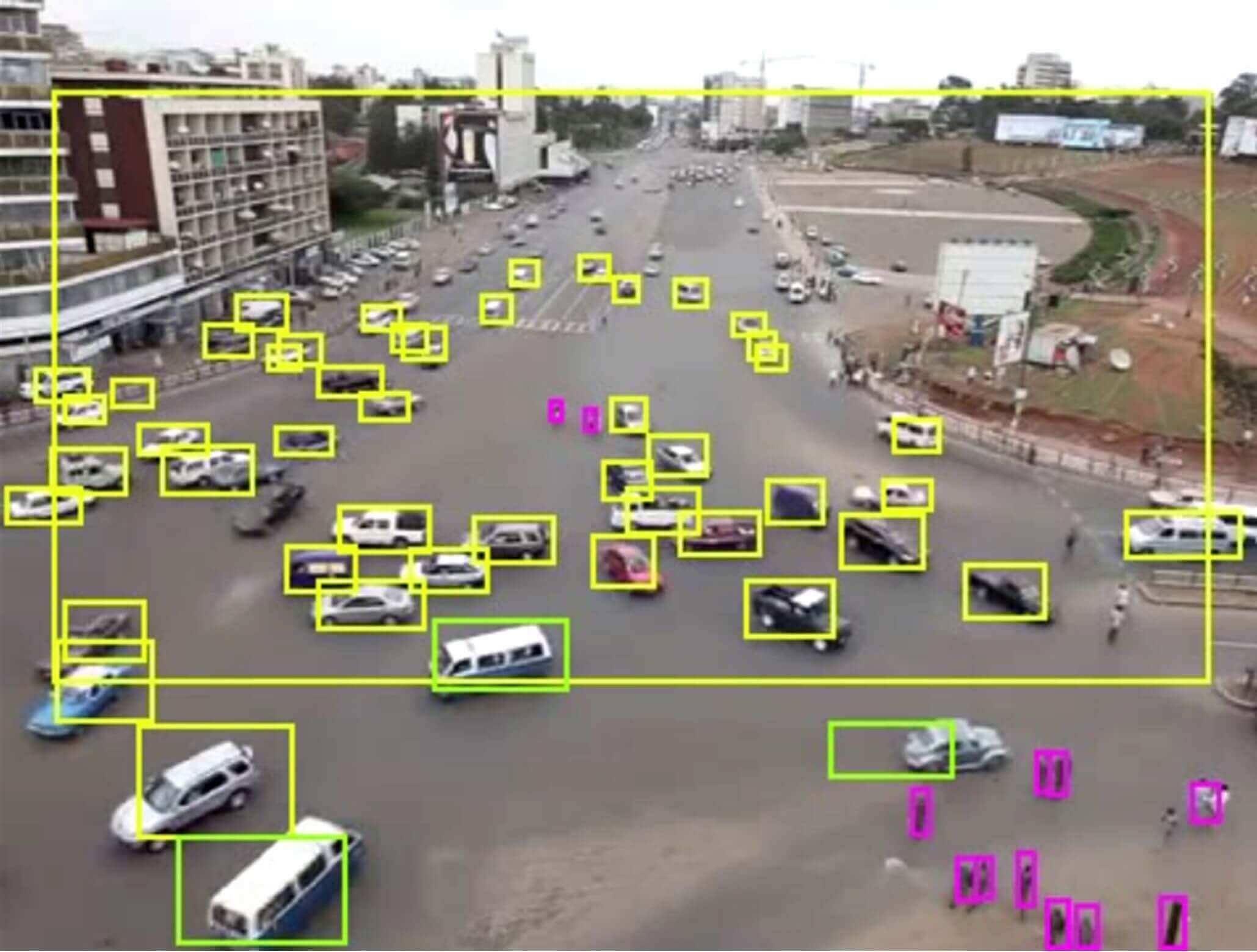}}
    \subfloat[]{\label{fig10d}
    \includegraphics[width=0.25\linewidth]{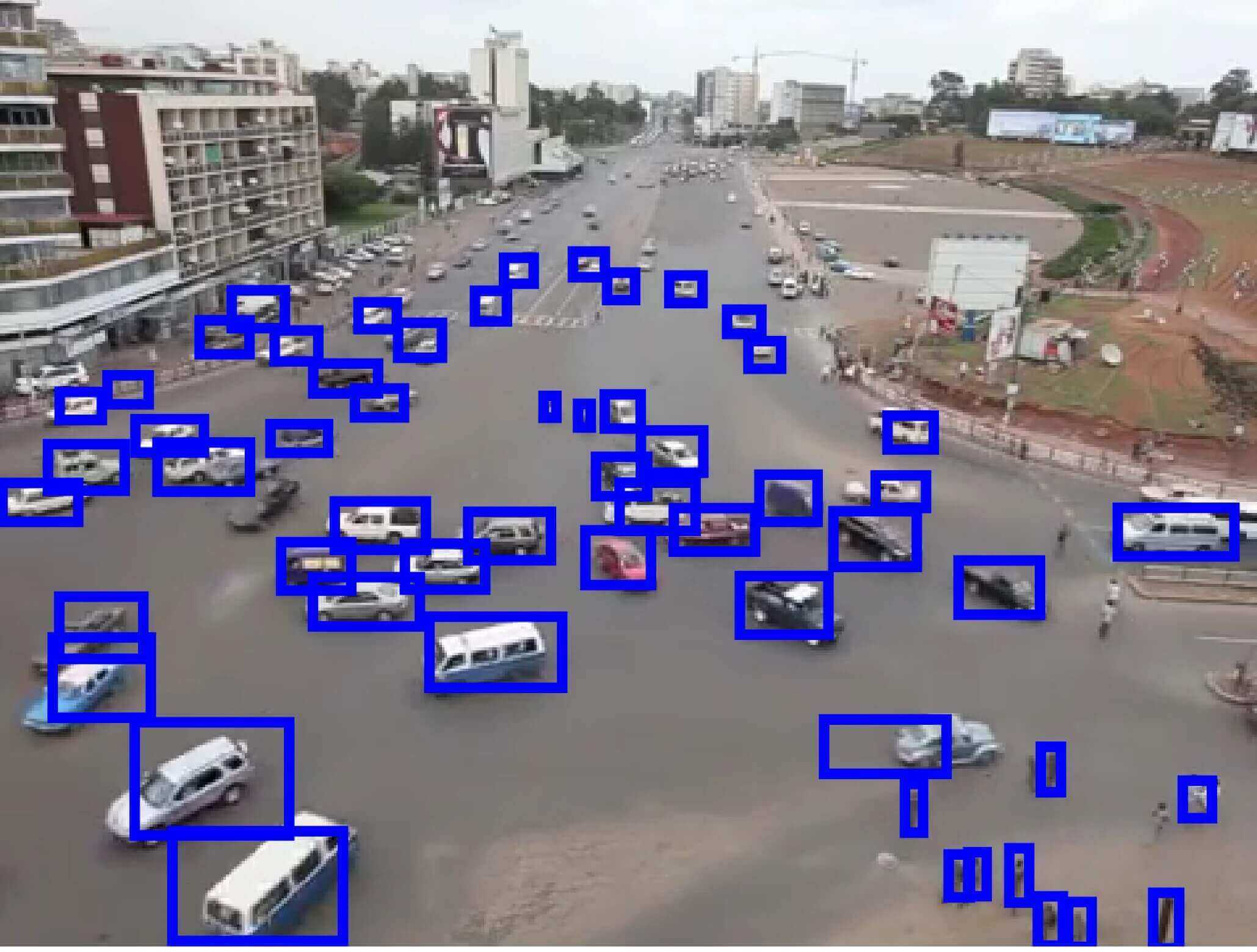}}
    \caption{\emph{YOLOv3} original detection results with threshold (a) $0.75$, (b) $0.50$, (c) $0.22$, and (d) the filtered result.}
    \label{fig10}
\end{figure}

As introduced above, we set a fixed ED threshold to determine whether a new ID will be assigned to a bounding box. However, the position and orientation of the camera would skew the ground slightly, and the size and velocity of objects would cause biases over different regions. Hence, we apply a perspective transformation to obtain a bird's-eye view video to deal with such biases by manually picking a rectangle $\overline{ABCD}$ as the reference points to perform a perspective transformation procedure, as shown in Fig.\ref{fig3}. The most frequent interaction behaviors occur in the red shaded area $\overline{CDEFG}$. Thus, we set the red area as the ROI and set the detection threshold of \emph{YOLOv3} as $0.22$ (Fig. \ref{fig10}). The big bounding boxes covering more than a quarter of the screen are removed. Furthermore, we set a filter on the overlapping ratio between the bounding boxes within one frame to avoid redundant detection from a low threshold. Figure \ref{fig10}\subref{fig10d} shows the filtered results, which are recorded in the form of sets of time series that contain the coordinates of all vehicles with the ID of each frame.

\section{Results and Analysis}\label{sec4}
In this section, we will present and analyze the experimental results tested on the Meskel Square Dataset and the NGSIM US Highway $101$ Dataset\cite{NGSIM}.

\subsection{Experimental Results}\label{subsecER}
\subsubsection{Gaussian Velocity Field}
Our focus is mainly on the vehicles that survive longer than $50$ frames with at least one surrounding vehicle, which allows us to find challenging and informative interaction patterns. The objects within the radius of $10$ m for each ego vehicle are considered and then the surrounding environment of each vehicle by extracting relative velocity is modeled using the coordinate transformation matrix. 
Human drivers' decision-making is more sensitive to ahead vehicles than vehicles on both sides, and the lateral speed is much lower than longitudinal speed for most scenarios. Therefore, the parameters of the Gaussian process are set as $A=1$, $\sigma_{x}=4$, and $\sigma_{y}=2$. 
The Gaussian velocity field is expressed as a set of matrices in the size of $11\times11\times2$, where $11$ represents the dimensions of $x$ and $y$ within the range of $[-10:10]$ m, and $2$ represents the velocity field in the $x$ and $y$ direction.

\subsubsection{Representation Learning} We build a deep autoencoder with a symmetric structure using fully connected layers. An $11$-layer deep autoencoder is trained for mapping Gaussian velocity field to hidden representations (i.e., the $6^{\mathrm{th}}$ hidden layer). More specifically, the number of hidden units in each encoder layer are set as $242$, $180$, $121$, $80$, $40$, and $24$, where $242$ and $24$ represent the input layer and the hidden representations, respectively.

\subsubsection{Traffic Primitives}
\begin{figure*}[t]
    \centering
    \includegraphics[width = 0.93\textwidth]{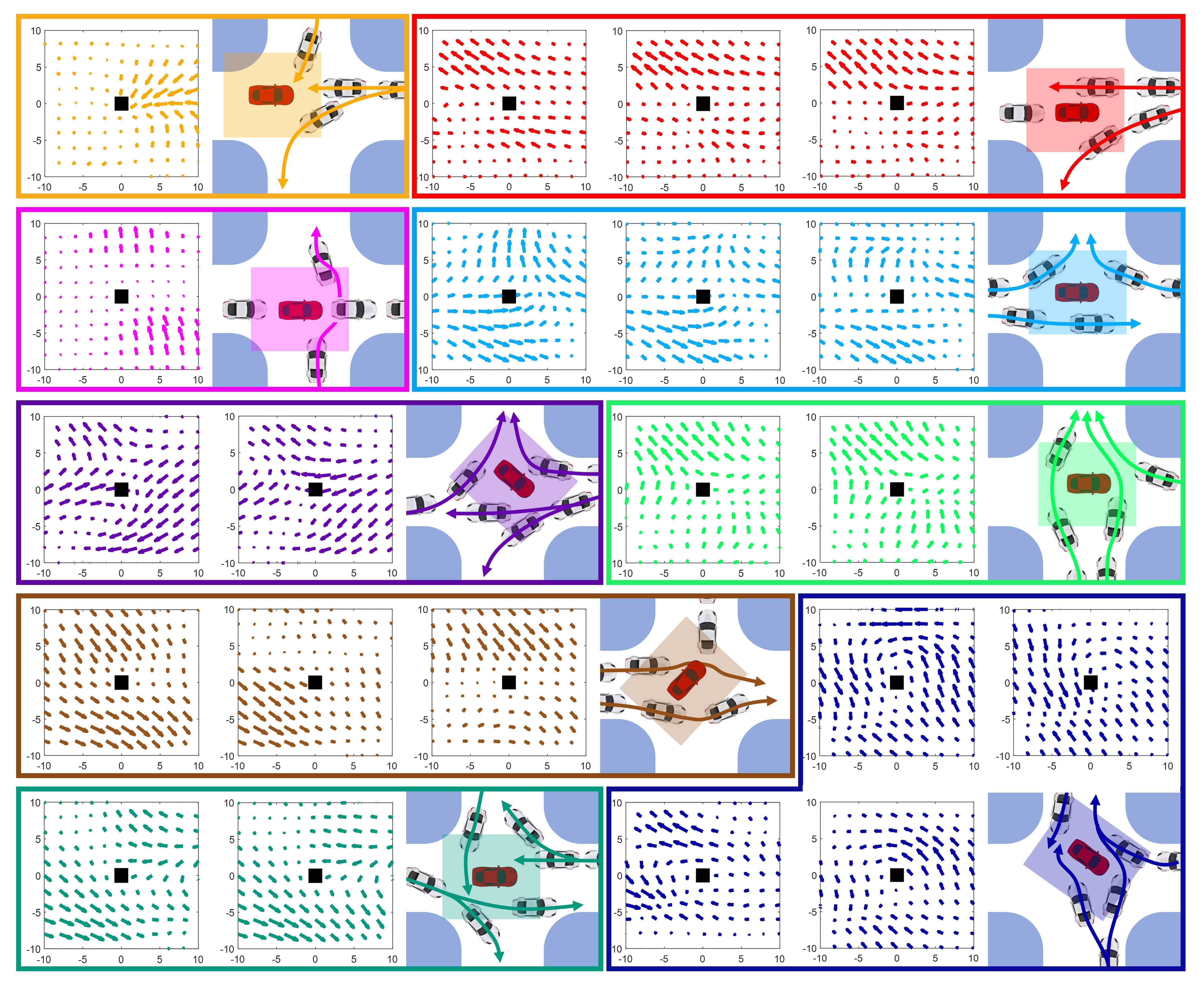}
    \caption{Nine traffic primitives in experimental results. The traffic primitives bounded in the same colored boxes are categorized into the same feature label, the left parts within each box are the representative velocity fields, and the right ones are their corresponding scenarios.}
    \label{fig11}
\end{figure*}

We use the latent features to represent relative velocity field for the surrounding environment and ego-vehicle states, and use the velocity $v_{ego}$ and acceleration $a_{ego}$ to depict the decisions of the ego vehicle. We focus on the first $1000$ typical vehicles and analyze the vehicles that are encountering complex traffic and struggling for a long duration. In addition, we neglect the vehicles that survive shorter than $50$ frames and without enough surrounding vehicles. 
With a total duration of $17279$ frames (which represents the summation of durations of all ego vehicles and is different from the video clip frames), the input dimension would be $17279\times26$, where $26$ elements are composed of $24$ hidden representations, $v_{ego}$, and $a_{ego}$. We train the HDP-HSMM for $1000$ epochs using \emph{pyhsmm}\cite{johnson2014bayesian}. Figure \ref{fig11} shows nine primitive interaction patterns extracted via our proposed framework. The learned traffic primitives convey scenarios of passing through among the opposite traffic flows, encountering with traffic flows from other directions, being stuck in the vertical traffic flows, and being cut off by other vehicles from the traffic flow that is being followed.

\begin{figure*}[ht]
    \centering
    \subfloat{\label{figa}
    \includegraphics[width=0.23\linewidth]{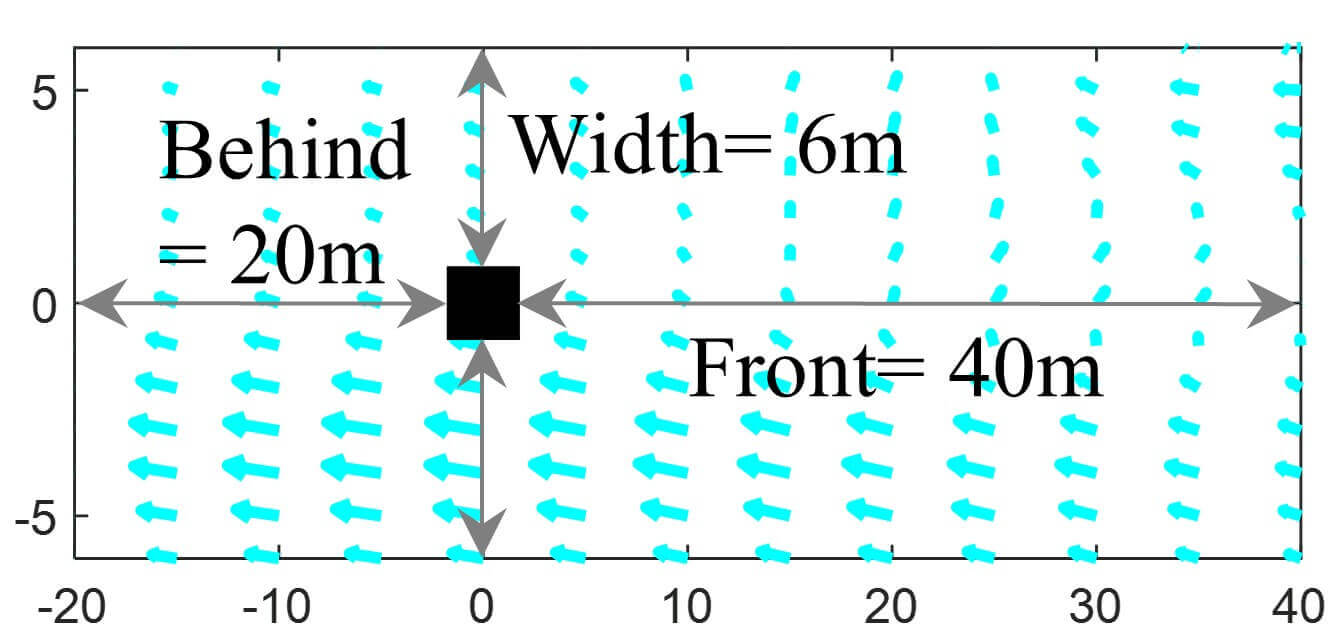}}\quad
    \subfloat{\label{figb}
    \includegraphics[width=0.23\linewidth]{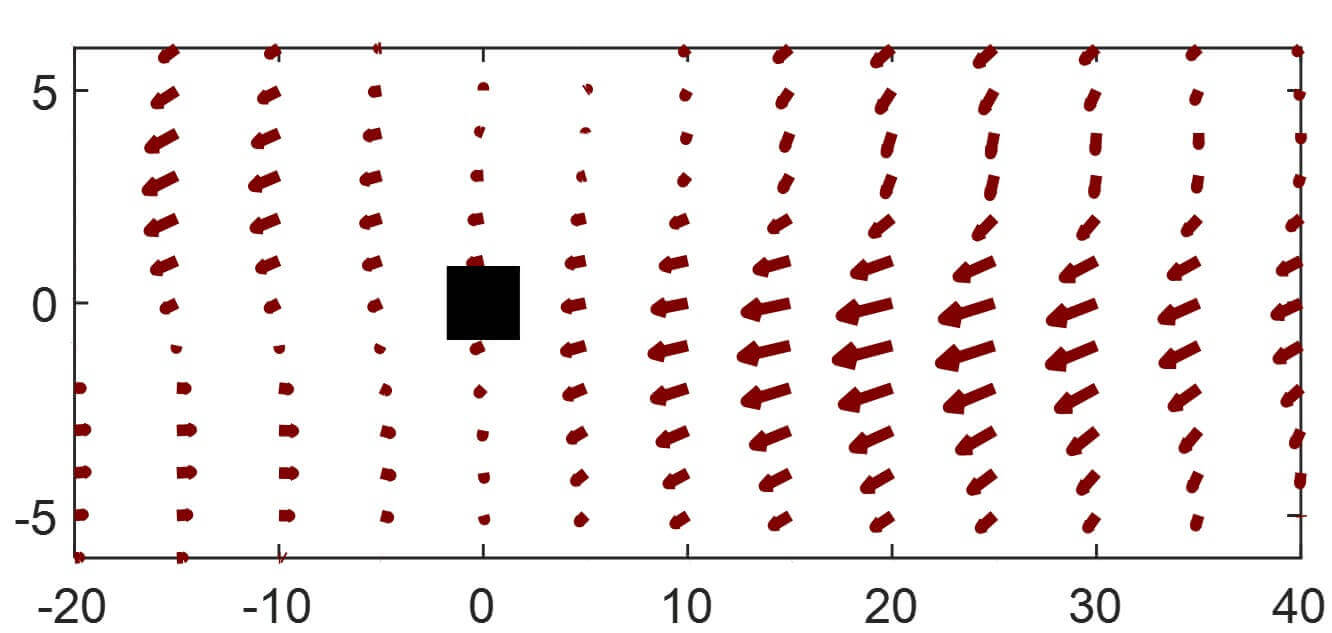}}\quad
    \subfloat{\label{figc}
    \includegraphics[width=0.23\linewidth]{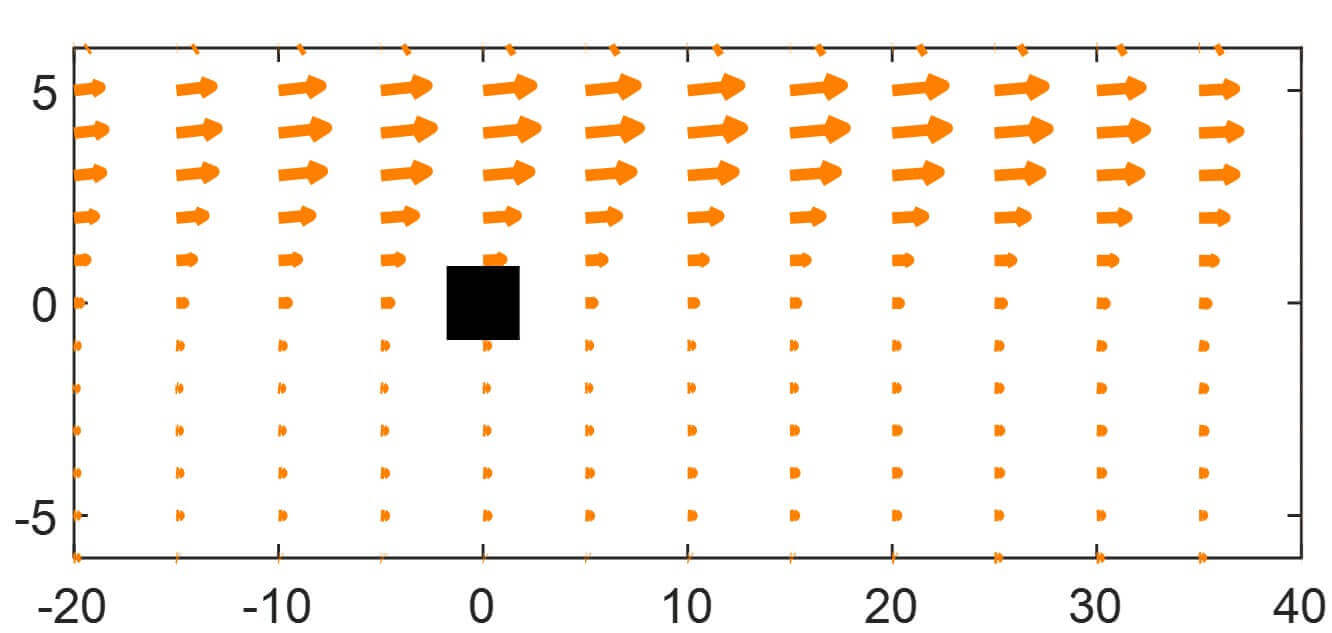}}\quad
    \subfloat{\label{figd}
    \includegraphics[width=0.23\linewidth]{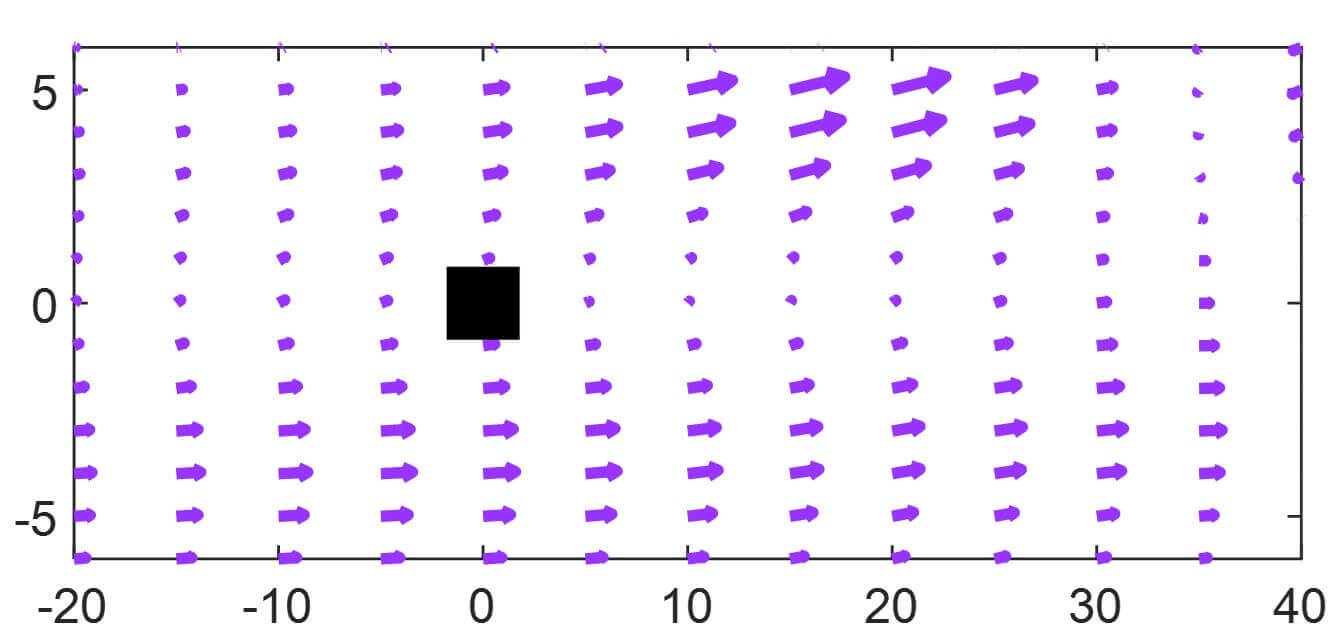}}\\
    \subfloat{\label{fige}
    \includegraphics[width=0.23\linewidth]{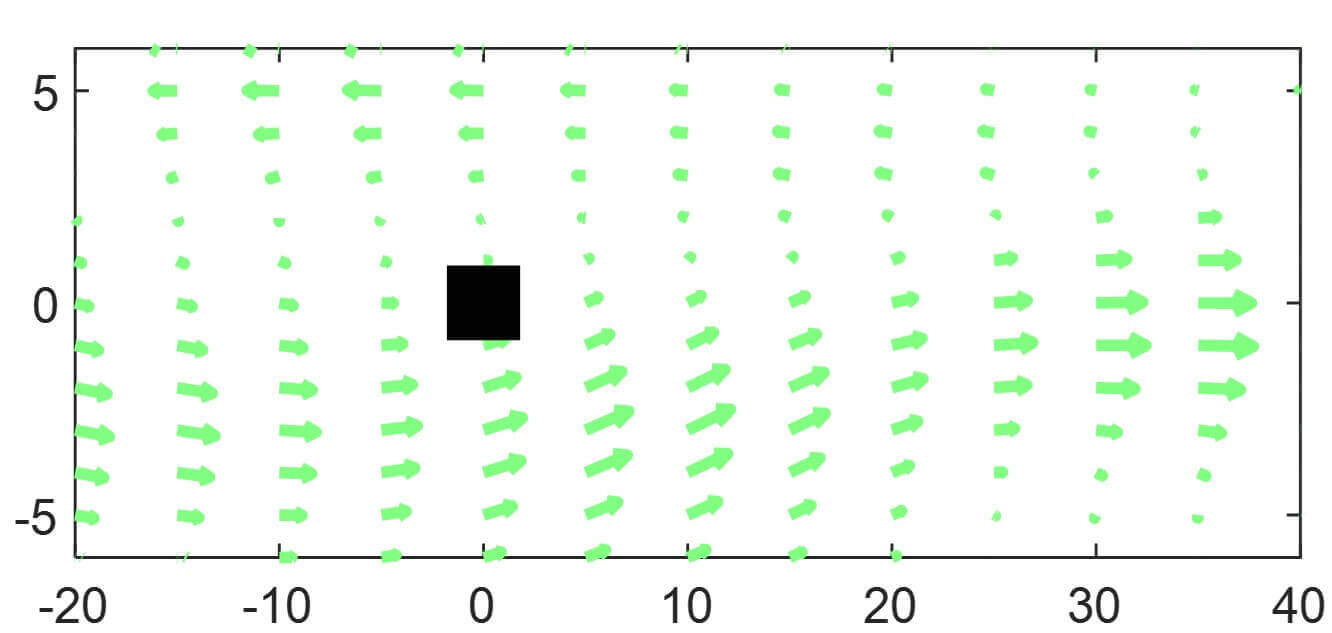}}\quad
    \subfloat{\label{figf}
    \includegraphics[width=0.23\linewidth]{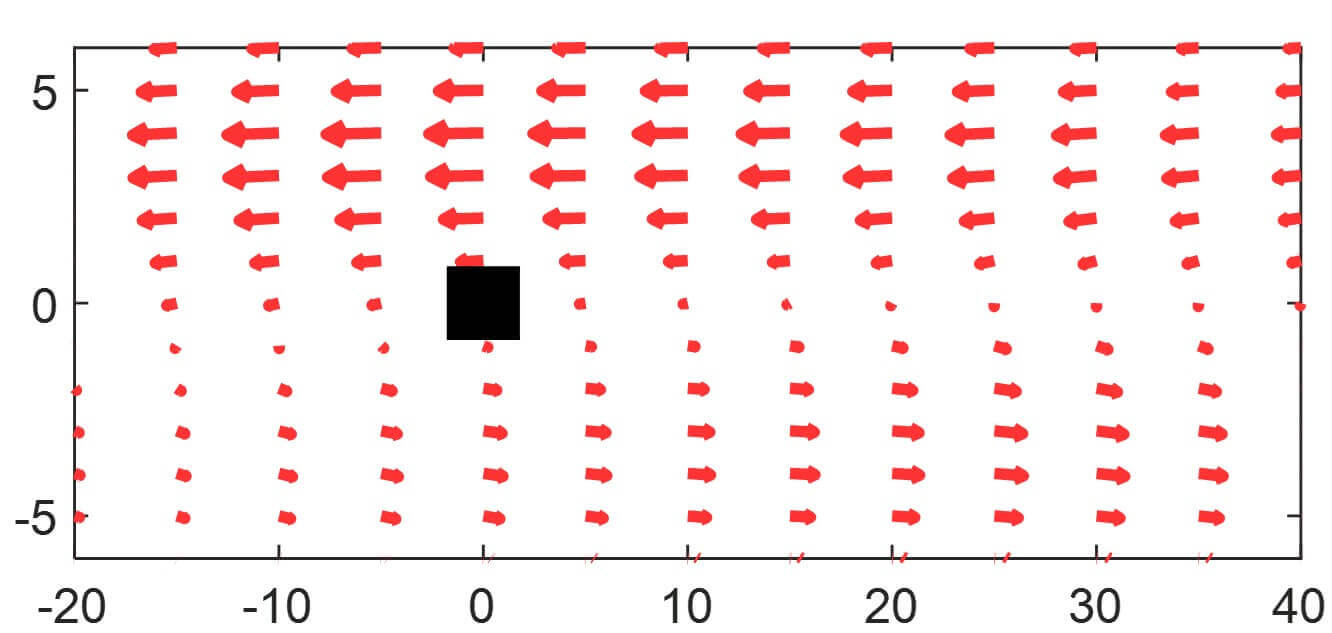}}\quad
    \subfloat{\label{figg}
    \includegraphics[width=0.23\linewidth]{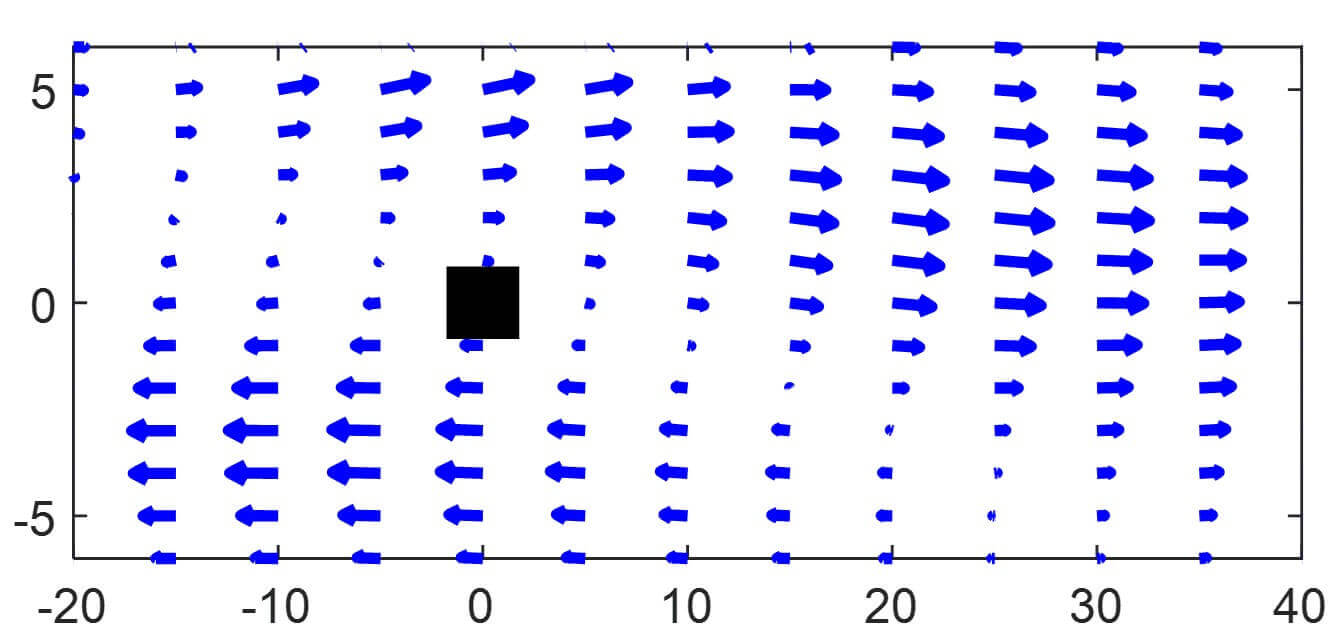}}\quad
    \subfloat{\label{figh}
    \includegraphics[width=0.23\linewidth]{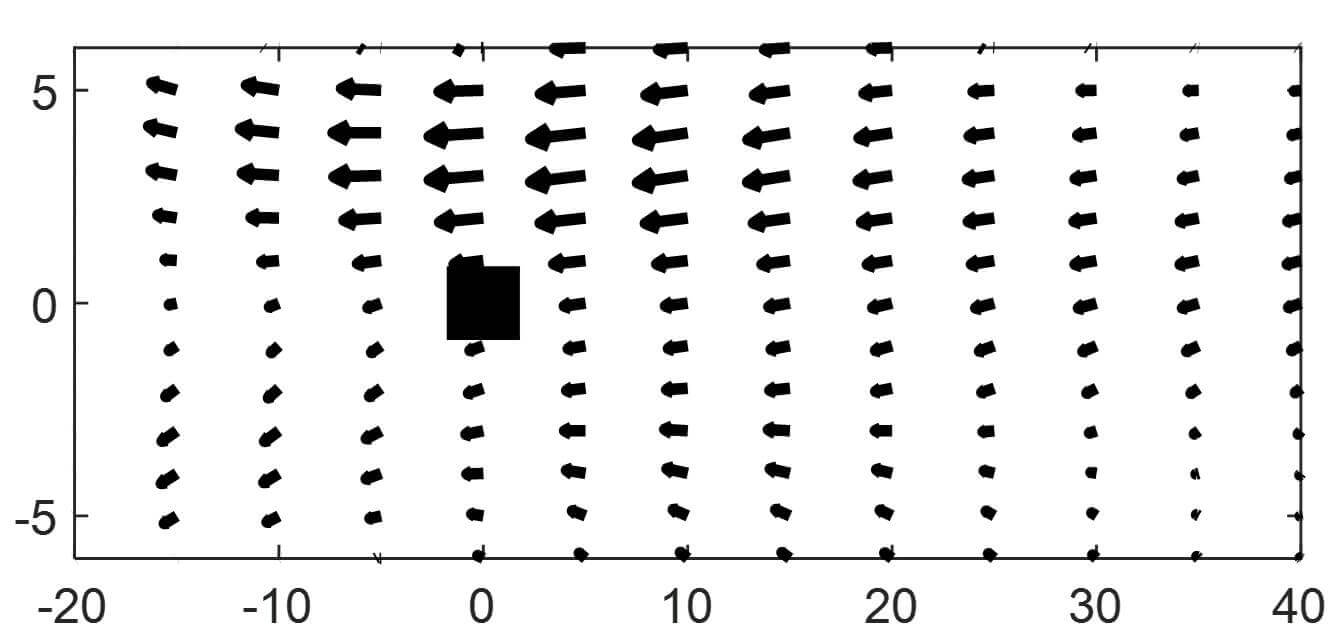}}\\
    \caption{The extracted eight traffic primitives from NGSIM datasets.}
    \label{fig_A2}
\end{figure*}

\subsection{Experiment on NGSIM Datasets}

Considering the differences between scenarios at intersections and on highways, we set the parameters of the Gaussian process as $A=1$, $\sigma_{x}=10$, and $\sigma_{y}=2$. 
In this experiment, we skip the first $39$ vehicles since the traffic is sparse at the very beginning of this dataset, and show the results of ID from $40-100$ with $14757$ frames. Eight representative interaction patterns on the highway are finally learned, as shown in Fig.\ref{fig_A2}. These interaction patterns include the fundamental highway scenarios, such as overtaking, being overtaken by the surrounding vehicles, and lane changing.

\section{Conclusion}\label{sec5}
In this paper, we demonstrated a flexible way to analyze complicated multi-vehicle interaction patterns based on traffic primitives from recorded traffic videos. A generic framework was developed, composing of three main modules: object tracking, representation learning, and interaction pattern clustering. This proposed framework can be extended to other scenarios with multi-agent involved. In the tracking module, we synthesized bounding boxes identification and optical flow representation to improve the position tracking and velocity estimation performance. In the representation learning module, we introduced the Gaussian velocity fields to model the cluttered scene where the number of vehicles is changing over time. Then, we learn the low-dimensional latent features of the velocity field by a deep autoencoder. Finally, in the interaction pattern clustering module, we input the combination of the velocity fields and the decisions of the ego vehicle into the Bayesian nonparametrics to extract and cluster the interaction patterns in the temporal space. 
Our experimental results on two different datasets show an appealing performance of the proposed framework for extracting semantically meaningful multi-vehicle interaction patterns. Such performance is desirable in the context of naturalistic driving analysis, especially in highly dynamic scenes. Furthermore, the resulting rich-information representation allows in-depth investigations for various autonomous driving applications, including reliable environment recognition, efficient scene understanding, and tractable safety evaluation.



\section*{ACKNOWLEDGMENT}
Toyota Research Institute (“TRI”) provided funds to assist the authors with their research but this article solely reflects the opinions and conclusions of its authors and not TRI or any other Toyota entity.
\bibliographystyle{IEEEtran}
\bibliography{references_ITSC2019}


\end{document}